\documentclass[letterpaper, 10 pt, conference]{IEEEtran}

\usepackage{caption}
\usepackage{amsmath,amssymb,amsthm,latexsym,float,epsfig,subfigure, listings, color,algorithm,algorithmic,url,multirow,csquotes,indentfirst,subcaption,comment}
\usepackage{rotating}
\usepackage{tablefootnote}
\usepackage[printonlyused]{acronym}
\usepackage[justification=justified,singlelinecheck=false]{caption}
\usepackage{booktabs}
\usepackage{longtable}  
\usepackage{svg}
\usepackage{bm} 
\usepackage{amsfonts}

\usepackage{xcolor}
\usepackage{graphicx}
\usepackage{makecell}
\usepackage{rotating}
\usepackage{threeparttable}

\usepackage{graphicx}
\usepackage{colortbl} 
\usepackage{rotating} 
\usepackage{multirow} 
\usepackage{adjustbox} 
\usepackage{tikz} 
\usepackage{array} 
\usepackage{float} 
\usepackage{afterpage}
\usepackage{placeins}
\usepackage{xurl} 
\usepackage{hyperref}

\begin{document}
\title{GaussianOcc3D: A Gaussian-Based Adaptive Multi-modal 3D Occupancy Prediction}

\author{A.~Enes~Doruk,
        and~Hasan~F.~Ates

\thanks{A. Enes Doruk and Hasan F. Ates are with the Department of Artificial Intelligence and Data Engineering, Ozyegin University, Istanbul, Turkey.}
}

\maketitle

\begin{abstract}
3D semantic occupancy prediction is a pivotal task in autonomous driving, providing a dense and fine-grained understanding of the surrounding environment, yet single-modality methods face trade-offs between camera semantics and LiDAR geometry. Existing multi-modal frameworks often struggle with modality heterogeneity, spatial misalignment, and the representation crisis—where voxels are computationally heavy and BEV alternatives are lossy. We present \textbf{GaussianOcc3D}, a multi-modal framework bridging camera and LiDAR through a memory-efficient, continuous 3D Gaussian representation. We introduce four modules: (1) \textbf{LiDAR Depth Feature Aggregation (LDFA)}, using depth-wise deformable sampling to lift sparse signals onto Gaussian primitives; (2) \textbf{Entropy-Based Feature Smoothing (EBFS)} to mitigate domain noise; (3) \textbf{Adaptive Camera-LiDAR Fusion (ACLF)} with uncertainty-aware reweighting for sensor reliability; and (4) a \textbf{Gauss-Mamba Head} leveraging Selective State Space Models for global context with linear complexity. Evaluations on Occ3D, SurroundOcc, and SemanticKITTI benchmarks demonstrate state-of-the-art performance, achieving mIoU scores of 49.4\%, 28.9\%, and 25.2\% respectively. GaussianOcc3D exhibits superior robustness across challenging rainy and nighttime conditions. \href{https://github.com/enesdoruk/GaussianOcc3D}{Code}
\end{abstract}


\section{Introduction}

\IEEEPARstart{T}{he}  3D semantic occupancy prediction is a pivotal task in the field of autonomous driving, providing a dense and fine-grained volumetric understanding of the surrounding environment~\cite{GenericADSurvey, openoccupancy2023}. Unlike traditional 3D object detection tasks that predominantly focus on specific classes of foreground objects using discrete bounding boxes, semantic occupancy prediction requires a comprehensive perception of the scene by predicting the geometry and semantics of both occupied and free space~\cite{GenericADSurvey, openoccupancy2023}. This capability is indispensable for navigating long-tail safety challenges, such as irregular roadway hazards, fallen rubble, or construction overhangs, which defy predefined rigid ontologies~\cite{GenericADSurvey}.

While recent advances have demonstrated significant progress in single-modality occupancy prediction. Camera-centric models, such as TPVFormer~\cite{yuan2023tpvformer} and GaussianFormer~\cite{huang2024gaussianformer}, leverage photometric richness for semantic understanding but suffer from the ill-posed nature of depth estimation from 2D images. Conversely, LiDAR-based solutions like JS3C-Net~\cite{cheng2021s3cnet} and LMSCNet~\cite{roldao2020lmscnet} provide precise geometric manifolds but lack the fine-grained semantic information required to distinguish visually similar classes.

\begin{figure*}[!t]
  \centering
  \includegraphics[width=0.85\textwidth]{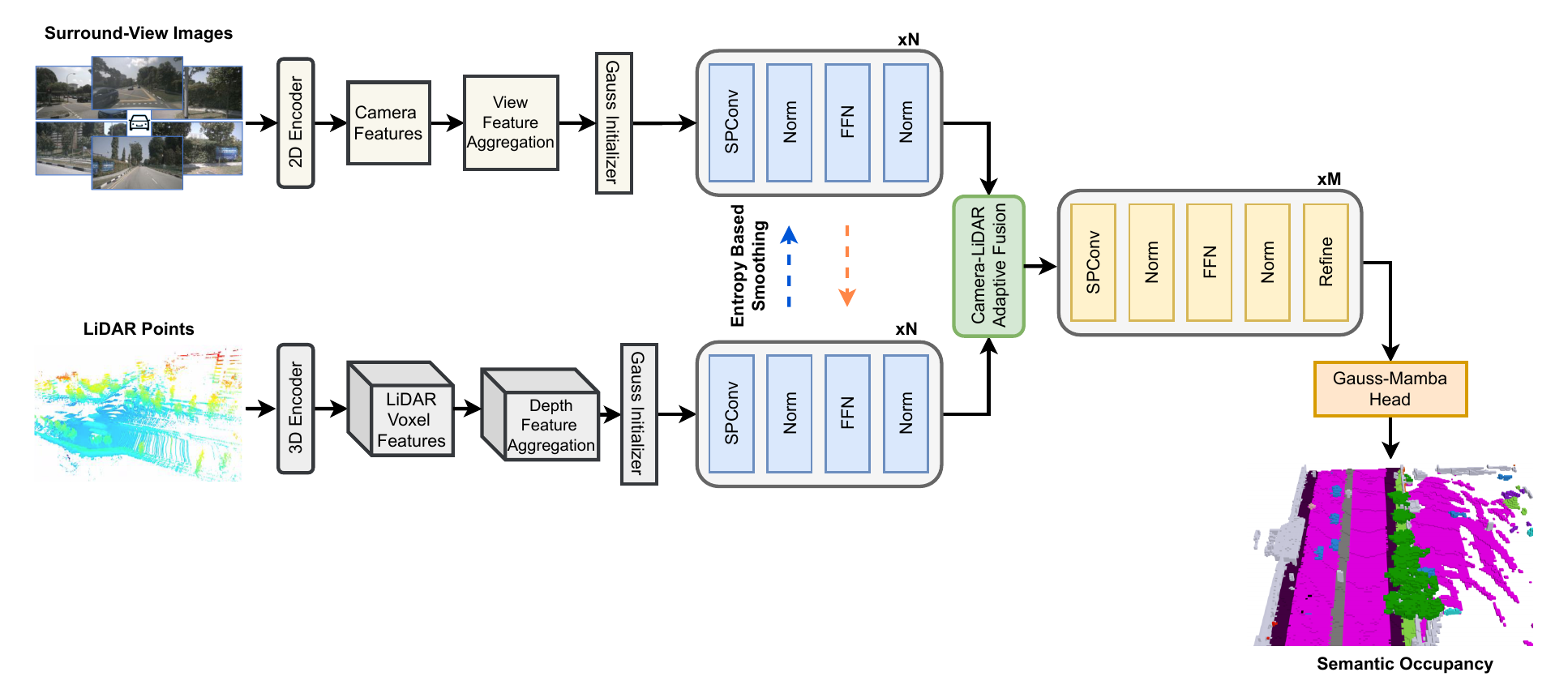}
  \vspace{-5pt}
  \caption{The overall architecture of the proposed GaussianOcc3D framework.}
  \label{fig:overview}
\end{figure*}

Multi-modal synergy seeks to resolve these ambiguities by leveraging the complementary strengths of both sensors. However, many existing multi-modal frameworks encounter significant difficulties in managing modality heterogeneity, spatial misalignment, and insufficient sensor interactions, which can result in the loss of critical semantic or geometric cues. These challenges are further compounded by a fundamental representation crisis in 3D perception: complex voxel-based models, such as OpenOccupancy~\cite{wang2023openoccupancy}, suffer from excessive computational demands due to the processing of empty space, while lightweight Bird's-Eye View (BEV) alternatives often rely on lossy compression that fails to accurately represent vertical structures and height information.

The emergence of 3D Gaussian Splatting~\cite{huang2024gaussianformer} has offered a continuous, memory-efficient alternative for environmental representation, notably explored in camera-centric models like GaussianFormer~\cite{huang2024gaussianformer, huang2024gaussianformer2}. However, vision-only Gaussian approaches often struggle with spurious geometry induced by depth ambiguity and overlook the robust geometric priors encoded in LiDAR data. To address these gaps, we propose GaussianOcc3D, a novel framework that bridges camera semantics and LiDAR geometry within a unified, shared Gaussian space. Our primary motivation is to overcome the sparsity of LiDAR data and the uncertainty of camera depth by natively lifting multi-modal signals onto adaptive Gaussian primitives. By utilizing anisotropic Gaussians that fit directly onto object surfaces rather than discretizing empty space, our model avoids the redundant processing inherent in voxel-based systems while ensuring geometric-semantic consistency. Our contributions are summarized as follows:

\begin{itemize}
    \item We propose GaussianOcc3D, a novel multi-modal occupancy prediction framework that bridges camera semantics and LiDAR geometry within a unified, memory-efficient 3D Gaussian representation.
    
    \item To address modality-specific challenges, we introduce the LiDAR Depth Feature Aggregation (LDFA) module for lifting sparse geometric signals via depth-wise deformable sampling, and the Entropy-Based Feature Smoothing (EBFS) module, which utilizes bidirectional cross-entropy to suppress inter-modal noise.
    
    \item We develop a dynamic Adaptive Camera-LiDAR Fusion (ACLF) mechanism featuring consistency-aware reweighting, alongside a Gauss-Mamba Head that leverages Selective State Space Models to capture long-range global context with linear complexity.
    
    \item Extensive evaluations on Occ3D, SurroundOcc, and SemanticKITTI benchmarks demonstrate state-of-the-art performance (e.g., 49.4\% mIoU on Occ3D) and superior robustness in adverse weather and low-light conditions.
\end{itemize}

\section{Related Work}

\textbf{3D Semantic Occupancy Prediction.} Recent research has transitioned from sparse 3D object detection toward dense semantic occupancy prediction to address long-tail safety challenges in autonomous driving \cite{openoccupancy2023}. Camera-based solutions  aim to obtain voxel-level semantic information from monocular or multi-view images. Early successes like FB-OCC \cite{li2023fbocc} utilized forward and backward view transformations, while OccDepth \cite{miao2023occdepth} integrated depth estimation to guide voxel assignments. More recent innovations include AdaOcc \cite{chen2024adaocc} for adaptive resolution. Despite these advances, vision-only methods remain limited by depth uncertainty. On the other hand, LiDAR-based  models like S3CNet \cite{cheng2021s3cnet} and LMSCNet \cite{roldao2020lmscnet} focus on 3D convolutions in voxel space. Subsequent developments include PointOcc \cite{zuo2023pointocc}, which uses a cylindrical tri-perspective view. While geometrically accurate, these models often struggle with semantic ambiguity in areas like road-sidewalk boundaries.

\textbf{Multi-Modal Fusion Strategies.} LiDAR and camera sensors is essential for capitalizing on the complementary strengths of geometric precision and semantic richness \cite{GenericADSurvey}. 

1) Projection-based fusion paradigm \cite{zhang2023occformer, li2024bevformer} combines image features with LiDAR points or projects point clouds into a range view to merge with photometric signals. 2) Feature-level fusion methods \cite{huang2021bevdet, huang2022bevdet4d} project both LiDAR and image data into a feature space, such as BEV or voxels, to learn cross-modal complementarities. 3) Attention-based fusion works \cite{li2024bevformer, yuan2023tpvformer} utilize LiDAR features or spatial queries to sample image features through cross-attention mechanisms. Recent works like OccMamba \cite{occmamba2024} have explored Mamba to handle large-scale occupancy grids. However, our method diverges by leveraging a Gaussian representation to expand the semantic perception field with aligned geometric abstractions, reducing errors from inaccurate calibrations.

\textbf{3D Gaussian Splatting.} Gaussian Splatting \cite{huang2024gaussianformer} has emerged as a memory-efficient alternative to traditional voxelization. These methods model the scene as a collection of anisotropic 3D Gaussians, concentrating computational resources on object surfaces rather than discretizing empty space, thus avoiding $O(N^3)$ complexity. Within occupancy prediction, GaussianFormer \cite{huang2024gaussianformer} first established a query-centric paradigm to predict Gaussian parameters. Building on this, GaussianFormer-2 \cite{huang2024gaussianformer2} proposed a probabilistic Gaussian superposition. While these are primarily camera-centric, our GaussianOcc3D leverages 3D Gaussians for both camera and LiDAR side, lifting multi-modal signals into a shared Gaussian space to maintain geometric and semantic consistency.

\section{Method}

\subsection{Framework Overview}

The architecture of GaussianOcc3D aims to unify multi-modal sensor streams within a continuous, memory-efficient 3D Gaussian representation. Given multi-view images $\mathcal{I} = \{I_i\}_{i=1}^{N_c}$ and LiDAR point clouds $\mathcal{P} = \{P_i\}_{i=1}^{N_p}$, where $P_i = (x_i, y_i, z_i, \eta_i)$ contains the 3D position and intensity, our goal is to predict a semantic occupancy grid $\hat{O} \in \mathcal{C}^{X \times Y \times Z}$. Unlike rigid voxel grids that suffer from cubic computational complexity, we model the scene as a set of 3D Gaussians $\mathcal{G} = \{G_i\}_{i=1}^{N_g}$. Each Gaussian $G_i$ is parameterized by its mean $m_i \in \mathbb{R}^3$, rotation $r_i \in \mathbb{R}^4$, scale $s_i \in \mathbb{R}^3$, opacity $\sigma_i \in [0, 1]$, and a semantic logit vector $c_i \in \mathbb{R}^{|C|}$ representing the class-wise scores.

The core principle of our representation is the universal approximation capability of Gaussian mixtures, which allows the model to adaptively fit anisotropic primitives to object surfaces. The semantic contribution of a single Gaussian evaluated at a location $x$ is defined as:
\begin{equation}
g(x; G) = \sigma \cdot \exp\left( -\frac{1}{2} (x - m)^T \Sigma^{-1} (x - m) \right) \cdot c,
\end{equation}
\vspace{-0.3cm}

where $\Sigma = RSS^TR^T$ represents the covariance matrix, with $R$ and $S$ being the rotation and scale matrices derived from $r$ and $s$. To determine the occupancy and semantics of a specific voxel, we employ a Gaussian-to-voxel splatting module that aggregates these contributions within a localized radius. The resulting semantic feature vector $\hat{o}(x)$ at the location $x$ is calculated as follows:
\begin{equation}
\hat{o}(x; \mathcal{G}) = \sum_{i=1}^{N_g(x)} g_i(x; m_i, s_i, r_i, \sigma_i, c_i).
\end{equation}

The final semantic label for each voxel is then determined by applying the $\text{argmax}(\cdot)$ operator over the aggregated logit vector $\hat{o}(x)$.


As illustrated in Fig. \ref{fig:overview}, the pipeline operates by first extracting 2D features via a camera backbone and 3D geometric features via a sparse LiDAR backbone. After Gaussian initialization, we employ an encoder consisting of SPConv, normalization, and an FFN to process these features. These features are then lifted into the shared Gaussian space through a multi-modal feature aggregation, utilizing query-based multi-view fusion for images \cite{huang2024gaussianformer} and our proposed LDFA module for sparse LiDAR signals. Before merging the modalities, we employ an entropy-based feature smoothing module to ensure local consistency within the Gaussian descriptors. Subsequently, the ACLF module adaptively fuses these representations to prioritize reliable sensory data. After fusion, the representations are refined through SPConv, normalization, and FFN layers to ensure feature consistency. Finally, the fused features are refined by the Gauss-Mamba Head to capture global context for the final occupancy prediction. The framework is trained end-to-end using a joint objective function:

\vspace{-0.3cm}
\begin{equation}
\mathcal{L}_{total} = \lambda_{ce} \mathcal{L}_{ce}(\hat{O}, \bar{O}) + \lambda_{lov} \mathcal{L}_{lov}(\hat{O}, \bar{O}),
\end{equation}
where $\mathcal{L}_{ce}$ is the cross-entropy loss and $\mathcal{L}_{lov}$ is the Lov\'{a}sz-softmax loss \cite{berman2018lovasz} optimized against ground truth labels $\bar{O}$.

\begin{figure}[!t]
  \centering
  \includegraphics[width=\columnwidth]{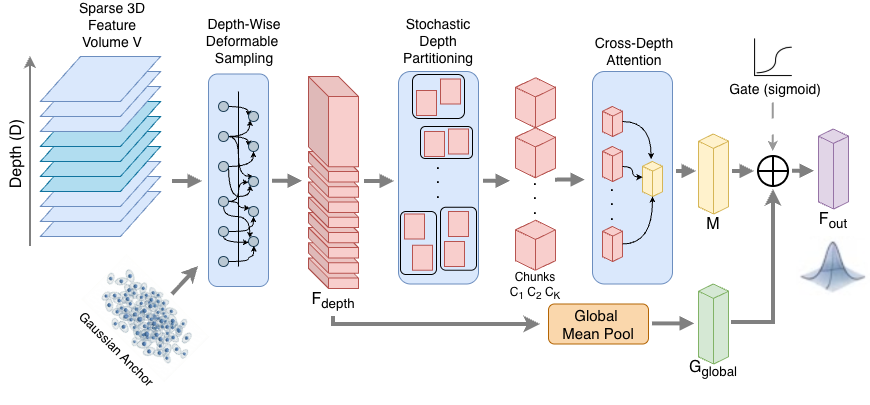}
  \vspace{-5pt}
  \caption{Architecture of the LiDAR Depth Feature Aggregation Module.}
  \label{fig:lcfa}
\end{figure}

\subsection{Multi-Modal Feature Aggregation}

Our architecture leverages a bifurcated processing stream to integrate dense photometric camera data and sparse geometric LiDAR data. For the camera branch, we adopt the multi-view aggregation paradigm of GaussianFormer \cite{huang2024gaussianformer}, treating 3D Gaussian anchors as spatial queries. These anchors are defined as learnable 3D centroids $\{a_i\}_{i=1}^{N_g}$ that serve as reference points to extract appearance features from surround-view images, ensuring photometric consistency across views.

The LiDAR branch utilizes a novel LiDAR Depth Feature Aggregation (LDFA) module. As illustrated in Fig. \ref{fig:lcfa}, we process the point cloud through a 3D sparse encoder to generate a voxelized feature volume $V \in R^{C \times D \times H \times W}$, where the vertical dimension is interpreted as a stack of $D$ depth planes. We employ a depth-wise deformable sampling strategy \cite{huang2024gaussianformer} where learnable offsets generate $P$ keypoints around each anchor. For an anchor $i$ at depth plane $d$, the stratified feature $f_{i,d}$ is computed as:

\vspace{-0.3cm}
\begin{equation}
    f_{i,d} = \sum_{k=1}^{P} w_{ik} \cdot P_d(u_{ik}),
\end{equation}
\vspace{-0.3cm}

where $w_{ik}$ is the learnable attention weight for the $k$-th sampling point, $u_{ik}$ represents the 2D projected coordinates of the keypoint on the $d$-th plane, and $P_d(\cdot)$ denotes the bilinear interpolation operator applied to the feature map of that specific depth level.

To address sparsity and enhance robustness, we implement stochastic depth partitioning, inspired by the chunk-based processing in DA-Mamba \cite{doruk2025mamba}. We divide the $D$ levels into $K$ disjoint chunks. During training, we apply a random permutation $\pi$ to the depth indices to prevent the model from overfitting to fixed spatial arrangements. The aggregated chunk representation $C_k$ is derived via mean pooling:

\vspace{-0.2cm}
\begin{equation}
    C_k = \frac{1}{|S_k|} \sum_{d \in S_k} F_{depth}^{(\pi(d))},
\end{equation}
\vspace{-0.25cm}

where $S_k$ is the set of indices belonging to the $k$-th chunk and $\pi(d)$ denotes the stochastic mapping of depth planes. 

A cross-depth attention mechanism then generates a modulation vector $M$ to identify valid surfaces by correlating chunk contexts. Finally, we re-introduce global context via a Gated Global Fusion mechanism, using a learnable soft gate $\alpha \in [0, 1]$ to balance modulated signals with the global mean feature $G_{global}$:
\begin{equation}
    \mathbf{F}_{out} = \alpha \cdot M + (1 - \alpha) \cdot G_{global}.
\end{equation}
This LDFA module ensures that Gaussian primitives are initialized with features that are geometrically grounded, significantly improving the fidelity of the 3D occupancy prediction.

\subsection{Entropy-Based Feature Smoothing}
\label{sec:smoothing}

The Entropy-Based Feature Smoothing (EBFS) module addresses the distributional misalignment that occurs when lifting camera and LiDAR features into a Gaussian space. Because images provide dense photometric texture while LiDAR provides sparse geometric manifolds, the resulting Gaussian feature descriptors often exhibit aliased distributions. This leads to the presence of characteristic rigid features that are non-transferable between modalities, where domain-specific sensor noise—such as visual glare or LiDAR multi-path reflections—corrupts the shared representation. EBFS quantifies uncertainty between streams via cross-entropy, dynamically rectifying the feature space to suppress these inconsistencies and artifacts before the fusion.

To ensure the framework learns intrinsically stable representations, we implement a stochastic execution strategy. Rather than applying smoothing to every layer, we randomly select a subset of layers to undergo refinement during each training iteration \cite{doruk2024transadapter}. This approach prevents the model from over-relying on the module to correct misalignments and serves as a regularization constraint that forces the encoder to produce robust camera and LiDAR features.

The core mechanism for quantifying sensor discrepancy is Cross-Entropy (CE). We interpret normalized feature vectors as probability distributions over the latent semantic space. Let $F_{C}$ and $F_{L}$ denote the  camera and LiDAR feature sets. We convert raw logits into probability distributions $P_{C}$ and $Q_{L}$ using Softmax scaled by temperature $\tau$:

\vspace{-0.5cm}
\begin{equation}
    P_{C} = \text{Softmax}\left(\frac{F_{C}}{\tau}\right), \quad Q_{L} = \text{Softmax}\left(\frac{F_{L}}{\tau}\right)
\end{equation}
\vspace{-0.4cm}

To refine the fused representation, we implement a bidirectional cross-entropy mechanism to quantify the alignment between modalities. We calculate two distinct entropy maps: camera-dominant entropy ($H_{C \rightarrow L}$), which measures LiDAR alignment with visual targets, and LiDAR-dominant entropy ($H_{L \rightarrow C}$), which identifies visual ambiguity relative to geometric references:

\vspace{-0.4cm}
\begin{equation}
    H_{m \rightarrow n} = - \sum_{c=1}^{C} P_{m}^{(c)} \cdot \log\left(Q_{n}^{(c)} + \xi \right)
\end{equation}
\vspace{-0.3cm}

where $(m,n) \in \{(C, L), (L, C)\}$. These maps are transformed into modulation weights $W_{C}$ and $W_{L}$ via exponential decay, normalized by the sum $\omega_{sum} = \sum \exp(-H) + \xi$. 

Finally, the features are updated via a residual connection with a learnable scaling parameter $\epsilon$ to adaptively tune the refinement intensity during training:

\vspace{-0.4cm}
\begin{equation}
    \tilde{F}_{i} = F_{i} + \epsilon \cdot W_{i}, \quad i \in \{C, L\}
\end{equation}
\vspace{-0.6cm}

This smoothing process suppresses noise in conflicting regions while reinforcing semantically consistent structures, leading to sharper boundaries in the final prediction.


\begin{figure}[!t]
  \centering
  \includegraphics[width=\columnwidth]{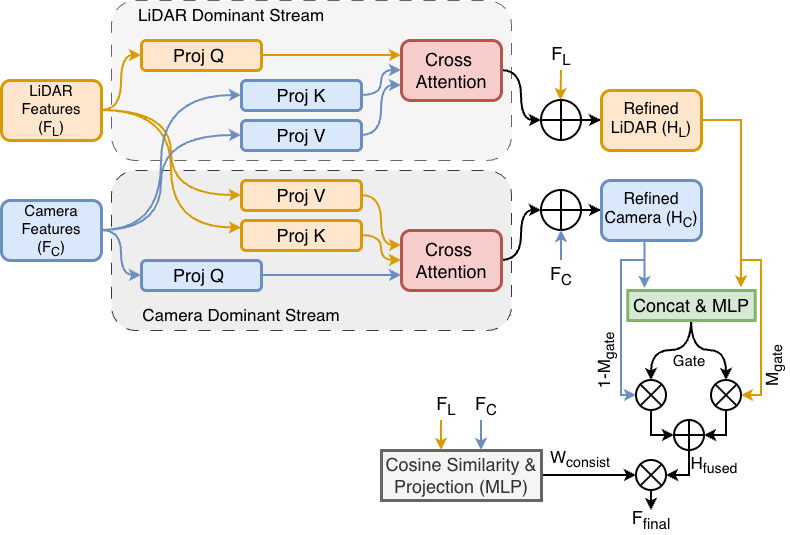}
  \vspace{-0.4cm}
  \caption{Architecture of the Adaptive Camera-LiDAR Fusion Module.}
  \label{fig:fusion}
\end{figure}

\definecolor{nbarrier}{RGB}{255, 120, 50}
\definecolor{nbicycle}{RGB}{255, 192, 203}
\definecolor{nbus}{RGB}{255, 255, 0}
\definecolor{ncar}{RGB}{0, 150, 245}
\definecolor{nconstruct}{RGB}{0, 255, 255}
\definecolor{nmotor}{RGB}{200, 180, 0}
\definecolor{npedestrian}{RGB}{255, 0, 0}
\definecolor{ntraffic}{RGB}{255, 240, 150}
\definecolor{ntrailer}{RGB}{135, 60, 0}
\definecolor{ntruck}{RGB}{160, 32, 240}
\definecolor{ndriveable}{RGB}{255, 0, 255}
\definecolor{nother}{RGB}{139, 137, 137}
\definecolor{nsidewalk}{RGB}{75, 0, 75}
\definecolor{nterrain}{RGB}{150, 240, 80}
\definecolor{nmanmade}{RGB}{213, 213, 213}
\definecolor{nvegetation}{RGB}{0, 175, 0}\definecolor{tan}{rgb}{0.82, 0.71, 0.55}

\begin{table*}[!ht]
  \centering
  \resizebox{0.85\linewidth}{!}{
    \begin{tabular}{c|c|c|cccccccccccccccc}
      \toprule
      Method & Modality & mIoU $\uparrow$ & 
      \rotatebox{90}{barrier} & 
      \rotatebox{90}{bicycle} & 
      \rotatebox{90}{bus} & 
      \rotatebox{90}{car} & 
      \rotatebox{90}{const. veh.} & 
      \rotatebox{90}{motorcycle} & 
      \rotatebox{90}{pedestrian} & 
      \rotatebox{90}{traffic cone} & 
      \rotatebox{90}{trailer} & 
      \rotatebox{90}{truck} & 
      \rotatebox{90}{drive. surf.} & 
      \rotatebox{90}{other flat} & 
      \rotatebox{90}{sidewalk} & 
      \rotatebox{90}{terrain} & 
      \rotatebox{90}{manmade} & 
      \rotatebox{90}{vegetation} \\
      & & & 
      \tikz \draw[fill=nbarrier,draw=nbarrier] (0,0) rectangle (0.2,0.2); & 
      \tikz \draw[fill=nbicycle,draw=nbicycle] (0,0) rectangle (0.2,0.2); & 
      \tikz \draw[fill=nbus,draw=nbus] (0,0) rectangle (0.2,0.2); & 
      \tikz \draw[fill=ncar,draw=ncar] (0,0) rectangle (0.2,0.2); & 
      \tikz \draw[fill=nconstruct,draw=nconstruct] (0,0) rectangle (0.2,0.2); & 
      \tikz \draw[fill=nmotor,draw=nmotor] (0,0) rectangle (0.2,0.2); & 
      \tikz \draw[fill=npedestrian,draw=npedestrian] (0,0) rectangle (0.2,0.2); & 
      \tikz \draw[fill=ntraffic,draw=ntraffic] (0,0) rectangle (0.2,0.2); & 
      \tikz \draw[fill=ntrailer,draw=ntrailer] (0,0) rectangle (0.2,0.2); & 
      \tikz \draw[fill=ntruck,draw=ntruck] (0,0) rectangle (0.2,0.2); & 
      \tikz \draw[fill=ndriveable,draw=ndriveable] (0,0) rectangle (0.2,0.2); & 
      \tikz \draw[fill=nother,draw=nother] (0,0) rectangle (0.2,0.2); & 
      \tikz \draw[fill=nsidewalk,draw=nsidewalk] (0,0) rectangle (0.2,0.2); & 
      \tikz \draw[fill=nterrain,draw=nterrain] (0,0) rectangle (0.2,0.2); & 
      \tikz \draw[fill=nmanmade,draw=nmanmade] (0,0) rectangle (0.2,0.2); & 
      \tikz \draw[fill=nvegetation,draw=nvegetation] (0,0) rectangle (0.2,0.2); \\
      \midrule
      MonoScene~\cite{cao2022monoscene} & C  & 7.3 & 4.0 & 0.4 & 8.0 & 8.0 & 2.9 & 0.3 & 1.2 & 0.7 & 4.0 & 4.4 & 27.7 & 5.2 & 15.1 & 11.3 & 9.0 & 14.9\\
      BEVFormer~\cite{li2024bevformer} & C & 16.8& 14.2& 6.6& 23.5& 28.3& 8.7& 10.8& 6.6& 4.1& 11.2& 17.8& 37.3& 18.0& 22.9& 22.2& 13.8& 22.2\\
      TPVFormer~\cite{yuan2023tpvformer} & C & 17.1& 16.0& 5.3& 23.9& 27.3& 9.8& 8.7& 7.1& 5.2& 11.0& 19.2& 38.9& 21.3& 24.3& 23.2& 11.7& 20.8\\
      OccFormer~\cite{zhang2023occformer} & C &  19.0& 18.7& 10.4& 23.9& 30.3& 10.3& 14.2& 13.6& 10.1& 12.5& 20.8& 38.8& 19.8& 24.2& 22.2& 13.5& 21.4\\
      SurroundOcc~\cite{surroundocc} & C &  20.3& 20.6& 11.7& 28.1& 30.9& 10.7& 15.1& 14.1& 12.1& 14.4& 22.3& 37.3& 23.7& 24.5& 22.8& 14.9& 21.9\\
      C-CONet~\cite{openoccupancy2023} & C &  18.4& 18.6& 10.0& 26.4& 27.4& 8.6& 15.7& 13.3& 9.7& 10.9& 20.2& 33.0& 20.7& 21.4& 21.8& 14.7& 21.3\\
      FB-Occ~\cite{li2023fbocc} & C & 19.6& 20.6& 11.3& 26.9& 29.8& 10.4& 13.6& 13.7& 11.4& 11.5& 20.6& 38.2& 21.5& 24.6& 22.7& 14.8& 21.6\\
      GaussianFormer~\cite{huang2024gaussianformer} & C &  19.1& 19.5& 11.3& 26.1& 29.8& 10.5& 13.8& 12.6& 8.7& 12.7& 21.6& 39.6& 23.3& 24.5& 23.0& 9.6& 19.1\\
      GaussianFormer-2~\cite{huang2024gaussianformer2} & C & 20.8& 21.4& 13.4& 28.5& 30.8& 10.9& 15.8& 13.6& 10.5& 14.0& 22.9& 40.6& 24.4& 26.1& 24.3& 13.8& 22.0 \\ 
      \midrule
      LMSCNet~\cite{roldao2020lmscnet} & L &  14.9& 13.1& 4.5& 14.7& 22.1& 12.6& 4.2& 7.2& 7.1& 12.2& 11.5& 26.3& 14.3& 21.1& 15.2& 18.5& 34.2\\
      L-CONet~\cite{openoccupancy2023} & L &  17.7& 19.2& 4.0& 15.1& 26.9& 6.2& 3.8& 6.8& 6.0& 14.1& 13.1& 39.7& 19.1& 24.0& 23.9& 25.1& 35.7\\
      \midrule
     M-CONet~\cite{openoccupancy2023} & C\&L & 24.7 & 24.8 & 13.0 & 31.6 & 34.8 & 14.6 & 18.0 & 20.0 & 14.7 & 20.0 & 26.6 & 39.2 & 22.8 & 26.1 & 26.0 & 26.0 & 37.1 \\
Co-Occ~\cite{coocc2024} & C\&L & \underline{27.1} & \underline{28.1} & 16.1 & \underline{34.0} & \underline{37.2} & 17.0 & 21.6 & 20.8 & \underline{15.9} & \underline{21.9} & 28.7 & \underline{42.3} & \textbf{25.4} & \textbf{29.1} & \underline{28.6} & 28.2 & 38.0 \\
OccFusion~\cite{occfusionming2024} & C\&L & 26.8 & 26.6 & \textbf{18.3} & 32.9 & 35.8 & \underline{19.3} & \underline{22.1} & \underline{24.4} & \textbf{17.7} & 21.4 & \underline{29.6} & 39.0 & 21.9 & 24.9 & 26.7 & 28.5 & 40.0 \\
GaussianFormer3D~\cite{gaussianformer3d2025} & C\&L & \underline{27.1} & 26.9 & 15.8 & 32.7 & 36.1 & 18.6 & 21.7 & 24.1 & 13.0 & 21.3 & 29.0 & 40.6 & 23.7 & \underline{27.3} & 28.2 & \underline{32.6} & \underline{42.3} \\
\rowcolor{black!10}
\textbf{GaussianOcc3D (ours)} & C\&L & \textbf{28.9} & \textbf{28.7} & \underline{16.8} & \textbf{34.8} & \textbf{38.5} & \textbf{19.8} & \textbf{23.1} & \textbf{25.7} & 13.9 & \textbf{22.7} & \textbf{30.9} & \textbf{43.3} & \underline{25.3} & \textbf{29.1} & \textbf{30.1} & \textbf{34.7} & \textbf{45.1} \\
      \bottomrule
    \end{tabular}
  }
  \vspace{-5pt}
  \caption{3D semantic occupancy prediction results on the SurroundOcc~\cite{surroundocc} validation set. The best and second-best are in bold and underlined, respectively. }
  \vspace{-5pt}
  \label{tab:surroundocc}
\end{table*}

\subsection{Adaptive Camera-LiDAR Fusion}
\label{sec:aclf}

To address the limitations of rigid sensor aggregation, we propose the Adaptive Camera-LiDAR Fusion (ACLF) module. This architecture treats fusion as a dynamic, context-aware process that explicitly models inter-modal consistency. Our approach is motivated by the fact that sensor reliability fluctuates per frame; for example, LiDAR may provide superior geometry in low-light conditions while the camera excels in capturing semantic boundaries during clear daylight, meaning no single modality is universally stable across all frames.

The process begins with dual-stream cross-attention refining modality-specific features $F_L$ and $F_C$ through cross-context. For the LiDAR-dominant stream, geometric queries $Q_L$ seek semantic texture from camera keys $K_C$ and values $V_C$:

\vspace{-0.2cm}
\begin{equation}
    H_L = F_L + \text{Softmax}\left(\frac{Q_L \cdot K_C^T}{\sqrt{d}}\right) V_C
\end{equation}

Symmetrically, the camera-dominant stream $H_C$ extracts structural context from the LiDAR features using queries $Q_C$.

To integrate these refined streams, we employ a soft gating mechanism. We project the concatenated features through a learned MLP to predict a fusion mask $M_{gate} \in [0, 1]$, allowing the network to adaptively weight sensor dominance:
\begin{equation}
    H_{fused} = M_{gate} \odot H_L + (1 - M_{gate}) \odot H_C
\end{equation}

Finally, to mitigate sensor conflict (e.g., visual hallucinations or LiDAR noise), we introduce a consistency-aware reweighting prior. This mechanism acts as a learned noise-suppression filter that utilizes the cosine similarity between projected camera and LiDAR latent spaces to quantify semantic consensus. Our motivation for using cosine similarity is to identify disagreements between sensors; a high similarity score indicates multi-modal agreement, while a low score signals potential hallucinations or domain-specific noise. By transforming this similarity into a channel-wise consistency gate $W_{consist}$, the model can adaptively suppress specific feature channels where sensors disagree—such as depth signals during a reflection—while preserving reliable signals. This adaptive filtering ensures that the features propagated to the final Gaussian representation are cross-verified, effectively eliminating ghost artifacts and improving overall robustness:
\begin{equation}
    F_{final} = H_{fused} \odot W_{consist}
\end{equation}

\subsection{Gauss-Mamba Head}
\label{sec:gauss_mamba}

The Gauss-Mamba Head models global spatial dependencies among 3D Gaussian primitives. While individual Gaussians capture local details, they lack the global awareness. To overcome the quadratic scaling limitations of traditional Transformer heads, we leverage the Selective State Space Model (Mamba) \cite{gu2023mamba}, which provides linear complexity and efficient long-range sequence modeling.

We treat the set of 3D Gaussians as a structured sequence by augmenting each primitive's feature with a positional encoding derived from its mean coordinates. To bridge the dimensionality gap, we implement a 3D-to-1D ordering strategy \cite{occmamba2024} that maps the spatial coordinates of the Gaussians into a continuous sequence while preserving local proximity. This ordered sequence is processed through a series of Mamba blocks using a selective scan mechanism to propagate information across the entire scene. By processing the Gaussians in this specific order, the head effectively captures environmental continuity—such as the linear extension of road surfaces or the vertical boundaries of urban structures—without the memory overhead of global self-attention. Following the global refinement, we apply a 1D-to-3D ordering to restore the spatial structure. These refined features are then utilized to predict the updated Gaussian parameters, including means, scales, rotations, opacities, and semantic descriptors. Finally, a Gaussian-to-voxel splatting module aggregates the contributions of these refined primitives into a dense semantic occupancy grid.

\begin{table*}[!ht]
  \centering
  \resizebox{0.85\linewidth}{!}{
    \begin{tabular}{c|c|c|ccccccccccccccccc}
      \toprule
      Method & Modality  & mIoU $\uparrow$ & 
      \rotatebox{90}{others} &
      \rotatebox{90}{barrier} & 
      \rotatebox{90}{bicycle} & 
      \rotatebox{90}{bus} & 
      \rotatebox{90}{car} & 
      \rotatebox{90}{const. veh.} & 
      \rotatebox{90}{motorcycle} & 
      \rotatebox{90}{pedestrian} & 
      \rotatebox{90}{traffic cone} & 
      \rotatebox{90}{trailer} & 
      \rotatebox{90}{truck} & 
      \rotatebox{90}{drive. surf.} & 
      \rotatebox{90}{other flat} & 
      \rotatebox{90}{sidewalk} & 
      \rotatebox{90}{terrain} & 
      \rotatebox{90}{manmade} & 
      \rotatebox{90}{vegetation} \\
      &  & & 
    \tikz \draw[fill=black,draw=black] (0,0) rectangle (0.2,0.2); &      
      \tikz \draw[fill=nbarrier,draw=nbarrier] (0,0) rectangle (0.2,0.2); & 
      \tikz \draw[fill=nbicycle,draw=nbicycle] (0,0) rectangle (0.2,0.2); & 
      \tikz \draw[fill=nbus,draw=nbus] (0,0) rectangle (0.2,0.2); & 
      \tikz \draw[fill=ncar,draw=ncar] (0,0) rectangle (0.2,0.2); & 
      \tikz \draw[fill=nconstruct,draw=nconstruct] (0,0) rectangle (0.2,0.2); & 
      \tikz \draw[fill=nmotor,draw=nmotor] (0,0) rectangle (0.2,0.2); & 
      \tikz \draw[fill=npedestrian,draw=npedestrian] (0,0) rectangle (0.2,0.2); & 
      \tikz \draw[fill=ntraffic,draw=ntraffic] (0,0) rectangle (0.2,0.2); & 
      \tikz \draw[fill=ntrailer,draw=ntrailer] (0,0) rectangle (0.2,0.2); & 
      \tikz \draw[fill=ntruck,draw=ntruck] (0,0) rectangle (0.2,0.2); & 
      \tikz \draw[fill=ndriveable,draw=ndriveable] (0,0) rectangle (0.2,0.2); & 
      \tikz \draw[fill=nother,draw=nother] (0,0) rectangle (0.2,0.2); & 
      \tikz \draw[fill=nsidewalk,draw=nsidewalk] (0,0) rectangle (0.2,0.2); & 
      \tikz \draw[fill=nterrain,draw=nterrain] (0,0) rectangle (0.2,0.2); & 
      \tikz \draw[fill=nmanmade,draw=nmanmade] (0,0) rectangle (0.2,0.2); & 
      \tikz \draw[fill=nvegetation,draw=nvegetation] (0,0) rectangle (0.2,0.2); \\
      \midrule
      MonoScene~\cite{cao2022monoscene} & C & 6.1& 1.8& 7.2& 4.3& 4.9& 9.4& 5.7& 4.0& 3.0& 5.9& 4.5& 7.2& 14.9& 6.3& 7.9& 7.4& 1.0& 7.7 \\
      BEVDet~\cite{huang2021bevdet} & C & 11.7& 2.1& 15.3& 0.0& 4.2& 13.0& 1.4& 0.0& 0.4& 0.13& 6.6& 6.7& 52.7& 19.0& 26.5& 21.8& 14.5& 15.3\\
      BEVFormer~\cite{li2024bevformer} & C & 23.7& 5.0& 38.8& 10.0& 34.4& 41.1& 13.2& 16.5& 18.2& 17.8& 18.7& 27.7& 49.0& 27.7& 29.1& 25.4& 15.4& 14.5\\
      TPVFormer~\cite{yuan2023tpvformer} & C & 28.3& 6.7& 39.2& 14.2& 41.5& 47.0& 19.2& 22.6& 17.9& 14.5& 30.2& 35.5& 56.2& 33.7& 35.7& 31.6& 20.0& 16.1\\
      RenderOcc~\cite{pan2024renderocc} & C & 26.1& 4.8& 31.7& 10.7& 27.7& 26.5& 13.9& 18.2& 17.7& 17.8& 21.2& 23.3& 63.2& 36.4& 46.2& 44.3& 19.6& 20.7\\
      GaussianFormer~\cite{huang2024gaussianformer} & C & 35.5& 8.8 & 40.9 & 23.3 & 42.9 & 49.7 & 19.2 & 24.8 & 24.4 & 22.5 & 29.4 & 35.3 & 79.0 & 36.9 & 46.6 & 48.2 & 38.8 & 33.1 \\
      BEVDet4D~\cite{huang2022bevdet4d} & C & 39.3 & 9.3& 47.1& 19.2& 41.5& 52.2& 27.2& 21.2& 23.3& 21.6& 35.8& 38.9& 82.5& 40.4& 53.8& 57.7& 49.9& 45.8 \\
      COTR~\cite{ma2024cotr} & C & 44.5& \underline{13.3}& \underline{52.1}& 32.0& 46.0& 55.6& \underline{32.6}& 32.8& 30.4& \underline{34.1}& 37.7& 41.8& \textbf{84.5}& \underline{46.2}& \textbf{57.6}& \underline{60.7}& 52.0& 46.3\\
      PanoOcc~\cite{openoccupancy2023} & C & 42.1& 11.7& 50.5& 29.6& 49.4& 55.5& 23.3& 33.3& 30.6& 31.0& 34.4& 42.6& \underline{83.3}& 44.2& 54.4& 56.0& 45.9& 40.4\\
      FB-Occ~\cite{li2023fbocc} & C & 42.1 & \textbf{14.3}& 49.7& 30.0& 46.6& 51.5& 29.3& 29.1& 29.4& 30.5& 35.0& 39.4& 83.1& \textbf{47.2}& 55.6& 59.9& 44.9& 39.6\\
      \midrule
      OccFusion~\cite{occfusionming2024} & C\&L & \underline{48.7} & 12.4 & 51.8 & \underline{33.0}& \underline{54.6}& 57.7 & \textbf{34.0} & \underline{43.0} & \underline{48.4} & \textbf{35.5}& \textbf{41.2}& 48.6& 83.0& 44.7& \underline{57.1}& 60.0& 62.5& 61.3 \\
      GaussianFormer3D~\cite{gaussianformer3d2025} & C\&L & 46.4 & 9.8 & 50.0 & 31.3 & 54.0 & \underline{59.4} & 28.1 & 36.2 & 46.2 & 26.7 & 40.2 &  \textbf{49.7} & 79.1 & 37.3 & 49.0 & 55.0 & \underline{69.1} & \underline{67.6}\\
      \rowcolor{lightgray!45}  \textbf{GaussianOcc3D (ours)} & C\&L & \textbf{49.4} & 12.5 & \textbf{54.2} & \textbf{34.5} & \textbf{56.4} & \textbf{59.7} & 32.4 & \textbf{43.7} & \textbf{48.6} & 32.7 & \underline{40.8} &  \underline{49.5} & 81.1 & 40.3 & 54.1 & \textbf{60.9} & \textbf{69.7} & \textbf{69.5}\\
      \bottomrule
    \end{tabular}
  }
  \vspace{-5pt}
  \caption{3D semantic occupancy prediction results on the Occ3D~\cite{tian2023occ3d} validation set. The best and second-best are in bold and underlined, respectively.}
  \vspace{-15pt}
  \label{tab:occ3d}
\end{table*}

\section{Experiments}
\subsection{Experimental Setup} 

\noindent \textbf{Benchmarks.} We conduct experiments on the NuScenes~\cite{nuscenes}, SurroundOcc~\cite{surroundocc}, Occ3D~\cite{occ3d}, and SemanticKITTI~\cite{behley2019semantickitti} benchmarks. NuScenes provides $1,000$ driving sequences annotated at $2\text{Hz}$, with SurroundOcc and Occ3D offering labels for $18$ categories across $700$ training and $150$ validation scenes. SurroundOcc discretizes a $[-50\text{m}, 50\text{m}]$ range at $0.5\text{m}$ resolution, while Occ3D targets a $[-40\text{m}, 40\text{m}]$ volume at $0.4\text{m}$ resolution including a visibility mask. For SemanticKITTI, we use sequences $00$--$10$ (excluding $08$) for training and $08$ for validation, employing a $256 \times 256 \times 32$ grid with a fine $0.2\text{m}$ voxel resolution and $19$ semantic classes. These datasets provide diverse environments and resolutions to evaluate the robustness of our Gaussian-based camera-LiDAR fusion framework.

\noindent \textbf{Evaluation Metrics.} Performance is assessed using the standard Intersection-over-Union ($\text{IoU}$) and mean IoU ($\text{mIoU}$) metrics, which measure the accuracy of geometric occupancy and semantic classification across all categories.

\noindent \textbf{Implementation Details.} For the camera branch, the input image resolutions are set to $900 \times 1600$ for NuScenes~\cite{nuscenes} and $1200 \times 1920$ for SemanticKITTI~\cite{behley2019semantickitti}. We utilize a ResNet101-DCN~\cite{resnet} backbone, initialized with a checkpoint pre-trained from FCOS3D~\cite{fcos3d}, along with a Feature Pyramid Network (FPN)~\cite{lin2017fpn} neck. In the LiDAR branch, we aggregate and voxelize the previous $10$ sweeps of point clouds, extracting mean features through a voxel feature encoder. The number of 3D Gaussian primitives is fixed at $25,600$. The model is trained using the AdamW optimizer with a weight decay of $0.01$. Learning rates are configured at $1 \times 10^{-4}$ for NuScenes and $2 \times 10^{-4}$ for SemanticKITTI, decaying via a cosine annealing schedule. Our framework is trained for $20$ epochs with a batch size of $1$ on Nvidia A100 GPUs.

\begin{figure}[!ht]
    \centering
    \includegraphics[width=0.35\textwidth]{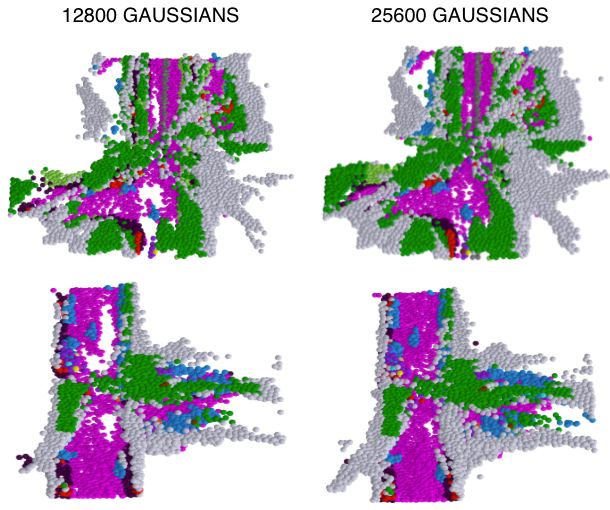} 
    \vspace{-5pt}
    \caption{Comparison of $12,800$ and $25,600$ Gaussian representations on the OpenOccupancy validation set. }
    \label{fig:gauss}
\end{figure}
\vspace{-0.6cm}

\subsection{Main Results}

We evaluate our model against SOTA methods across three benchmarks: SurroundOcc \cite{surroundocc}, Occ3D \cite{occ3d}, and SemanticKITTI \cite{semantickitti}. 

\begin{figure*}[t]
    \centering
    \includegraphics[width=0.9\textwidth]{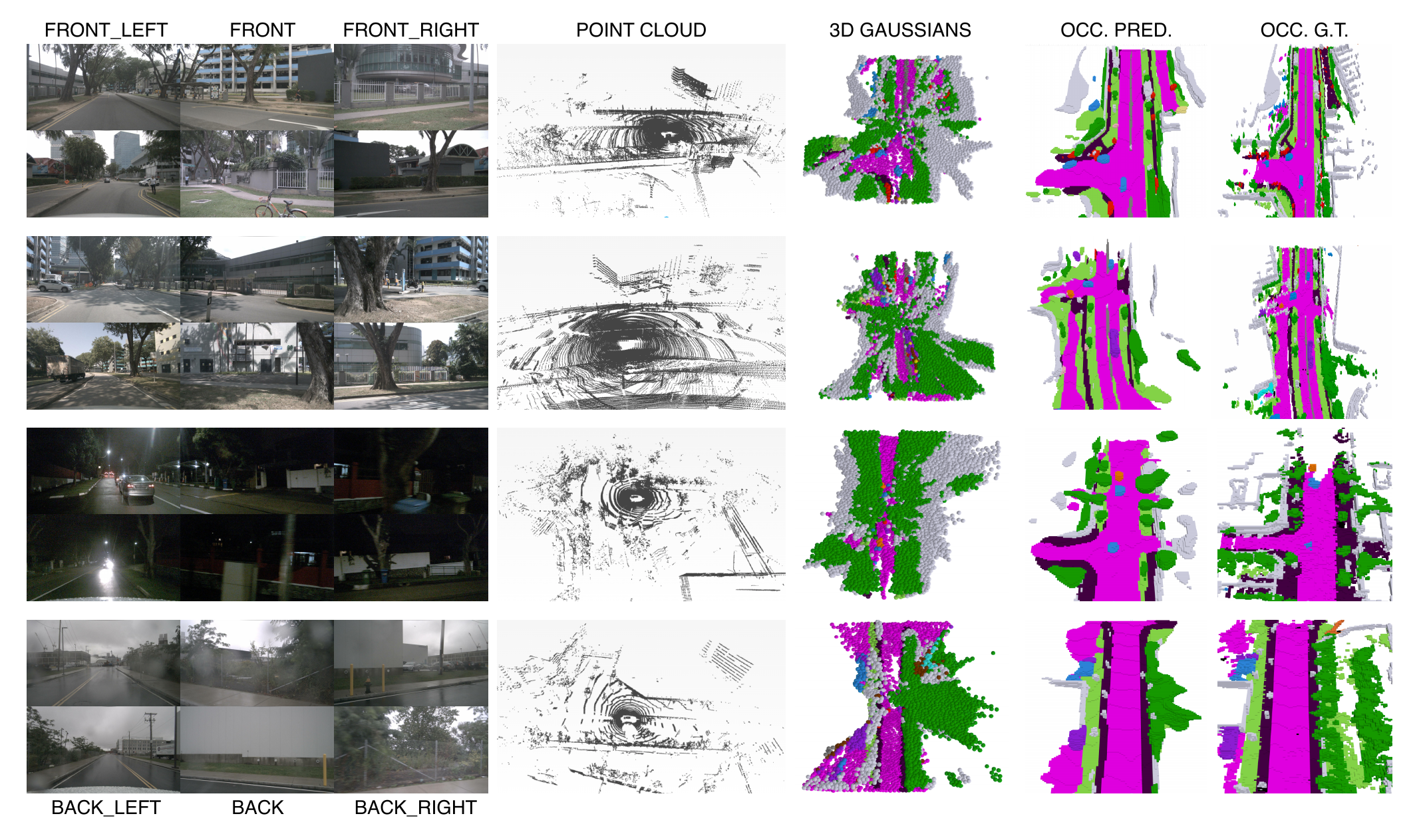} 
    \vspace{-5pt}
    \caption{Qualitative result on the SurroundOcc \cite{surroundocc} validation set.}
    \label{fig:qual}
\end{figure*}

In the SurroundOcc benchmark (Table~\ref{tab:surroundocc}), our framework achieves a SOTA mIoU of 28.9\%. While single-modality methods like BEVFormer and SurroundOcc demonstrate the limitations of relying on camera alone, our approach significantly outperforms existing multi-modal architectures. Specifically, we surpass M-CONet by 4.2\% and maintain a clear margin over fusion baselines such as Co-Occ and GaussianFormer3D, both of which reach 27.1\%. Our model shows superior performance in challenging categories such as pedestrian (23.1\%) and traffic cone (25.7\%), where precise alignment enables fine-scale detection. These improvements are qualitatively evidenced in Fig.~\ref{fig:qual}, where our model generates sharper occupancy boundaries and more accurate predictions for small-scale objects.

\begin{table}[!h]
\centering
\fontsize{8pt}{10pt}\selectfont
\begin{threeparttable}
\begin{tabular}{l|c|c} 
\toprule
Method & Input Modality & mIoU \\
\midrule
MonoScene~\cite{cao2022monoscene} & C & 11.1 \\
SurroundOcc~\cite{surroundocc} & C & 11.9 \\
OccFormer~\cite{zhang2023occformer} & C & 12.3 \\
RenderOcc~\cite{pan2024renderocc} & C & 12.8 \\
\cmidrule{1-3}
LMSCNet~\cite{roldao2020lmscnet} & L & 17.0 \\
JS3C-Net~\cite{cheng2021s3cnet} & L & 23.8 \\
SSC-RS~\cite{mei2023ssc} & L & 24.2 \\
\cmidrule{1-3}
Co-Occ~\cite{coocc2024} & C\&L & 24.4 \\
M-CONet~\cite{openoccupancy2023} & C\&L & 20.4 \\
OccMamba~\cite{occmamba2024} & C\&L & \underline{24.6} \\
\rowcolor{black!10}

\textbf{GaussianOcc3D (ours)} & C\&L & \textbf{25.2} \\

\bottomrule
\end{tabular}
\vspace{-5pt}
\caption{Performance on the SemanticKITTI test set. The best and second-best are in bold and underlined, respectively.}
\label{table_semantickitti}
\end{threeparttable}
\end{table}
\vspace{-0.3cm}

On the \textbf{Occ3D \cite{occ3d}} dataset, summarized in Table~\ref{tab:occ3d}, our model reaches a peak performance of 49.4\% mIoU. This establishes a new SOTA for multi-modal occupancy prediction, outperforming camera-only methods like FB-Occ \cite{li2023fbocc} and PanoOcc \cite{openoccupancy2023}. Compared to the leading multi-modal baseline OccFusion (48.7\%), our framework achieves significant gains in structural categories such as barrier (54.2\%) and bus (56.4\%). The results indicate that our Gaussian-based representation successfully handles the high-resolution requirements of Occ3D while suppressing the semantic noise often found in dense voxel-based fusion methods.

Finally, evaluation on the \textbf{SemanticKITTI} test set, presented in Table~\ref{table_semantickitti}, confirms the robustness of our approach with a leading mIoU of \textbf{25.2\%}. In this benchmark, our model outperforms specialized multi-modal architectures including OccMamba \cite{occmamba2024} (24.6\%), Co-Occ \cite{coocc2024} (24.4\%), and M-CONet \cite{openoccupancy2023} (20.4\%). The performance advantage is particularly meaningful given that SemanticKITTI \cite{semantickitti} features sparser LiDAR data compared to nuScenes \cite{nuscenes}; our framework's ability to exceed the performance of single-modality LiDAR baselines like SSC-RS \cite{mei2023ssc} (24.2\%) highlights the effectiveness of our lifting strategy in maintaining geometric fidelity across different sensor configurations and spatial resolutions.

\begin{table}[h]
\centering
\scalebox{1}{
\begin{tabular}{ccc|c}
\toprule
Addition & Concatenation & ACLF & mIoU         \\ 
\midrule
     \checkmark&  & & 28.0  \\ 
     & \checkmark&  & \underline{28.2}  \\ 
     \rowcolor{black!10}
     & &  \checkmark & \textbf{28.9}  \\ 
\bottomrule
\end{tabular}
}
\vspace{-5pt}
\caption{Performance analysis of different fusion techniques on SurroundOcc~\cite{surroundocc} validation set. The best and second-best are in bold and underlined, respectively.}
\label{table:fusion}
\end{table}

\vspace{-0.4cm}
\subsection{Ablation Studies}


We evaluate the analyses on SurroundOcc \cite{surroundocc} dataset.

\begin{table}[h]
    \centering
    \resizebox{0.8\linewidth}{!}{
     \begin{tabular}{c|c|cc}
        \toprule
        &  & \multicolumn{2}{c}{mIoU$\uparrow$}         \\
        \multirow{-2}{*}{Method} & \multirow{-2}{*}{Modality}  &  Rainy & Night\\ 
        \midrule
        GaussianFormer~\cite{huang2024gaussianformer} & C   &  18.0  & 9.3 \\
        GaussianFormer3D~\cite{gaussianformer3d2025} & C\&L  &  25.2  & \underline{15.5} \\
        OccFusion~\cite{occfusionming2024} & C\&L   &  \underline{26.5}  & 15.2 \\
        \rowcolor{black!10}
        \textbf{GaussianOcc3D (ours)} & C\&L  & \textbf{27.1}   & \textbf{15.9} \\
        \bottomrule
      \end{tabular}
    }
    \vspace{-5pt}
    \caption{3D semantic occupancy prediction results on SurroundOcc~\cite{surroundocc} validation set for rainy and night weather. The best and second-best are in bold and underlined, respectively.} 
    \label{tab:weather}
\end{table}

\vspace{-0.3cm}
\textbf{Fusion Strategy.} We compare our adaptive fusion approach against standard techniques in Table~\ref{table:fusion}. Traditional element-wise addition and concatenation yield 28.0\% and 28.2\% mIoU. Our proposed method outperforms these rigid aggregation mechanisms with 28.9\% mIoU, demonstrating that dynamic, consistency-aware weighting is essential for resolving multi-sensor conflicts and preserving valid geometric signals.

\textbf{Model Complexity.} Table~\ref{table:complexity} analyzes the trade-off between performance and efficiency. While the camera-only GaussianFormer \cite{huang2024gaussianformer} is lighter, our GaussianOcc3D maintains a competitive footprint with \underline{68.1M} parameters and \underline{427ms} latency, significantly outperforming the multi-modal OccMamba \cite{occmamba2024} in both speed and memory efficiency. This indicates that our architecture provides a superior balance of high-fidelity 3D reconstruction and computational feasibility.

\vspace{-0.2cm}
\begin{table}[h]
  \centering
  \resizebox{\linewidth}{!}{
    \begin{tabular}{l|c|c|c|c}
      \toprule
      Method & Mod. & \makecell{Query\\Form} & \makecell{Query\\Number} & mIoU $\uparrow$ \\
      \midrule
      & &  & 6400  & 19.9 \\
      & &  & 12800   & 19.9 \\
      \multirow{-3}{*}{GaussianFormer-2~\cite{huang2024gaussianformer2}} & \multirow{-3}{*}{C} & \multirow{-3}{*}{\makecell{3D\\Gaussian}} & 25600  & 20.3 \\
      \midrule
      M-CONet~\cite{openoccupancy2023} & C\&L & 3D Voxel & 100$\times$100$\times$8 & 24.7 \\
      \cmidrule(lr){1-5}
      Co-Occ~\cite{coocc2024} & C\&L & 3D Voxel & 100$\times$100$\times$8 & \underline{27.1} \\
      \cmidrule(lr){1-5}
      & & & 12800  & 24.2 \\
      \multirow{-2}{*}{GaussianFormer3D~\cite{gaussianformer3d2025}} & \multirow{-2}{*}{C\&L} & \multirow{-2}{*}{\makecell{3D\\Gaussian}} & 25600  & \underline{27.1} \\
      \cmidrule(lr){1-5}
      \rowcolor{black!10}
      \rowcolor{black!10}
      & & & 12800  & 26.1 \\
      \rowcolor{black!10}
      \multirow{-2}{*}{\textbf{GaussianOcc3D (ours)}} & \multirow{-2}{*}{C\&L} & \multirow{-2}{*}{\makecell{3D\\Gaussian}} & 25600  & \textbf{28.9} \\
      \bottomrule
    \end{tabular}
  }
  \vspace{-5pt}
  \caption{Efficiency comparison of different methods on SurroundOcc~\cite{surroundocc} validation set. The best and second-best are in bold and underlined, respectively.}
  \label{tab:effc}
\end{table}

\vspace{0.2cm}
\textbf{Effect of Gaussian Density.} The relationship between the number of primitives and accuracy is detailed in Table~\ref{tab:effc}. Increasing the gaussians from 12,800 to 25,600 gaussians leads to a performance gain, reaching 28.9\% mIoU. Compared to voxel-based methods like Co-Occ \cite{coocc2024} using fixed grids, our Gaussian-based representation achieves higher mIoU with more efficient spatial modeling, as visualized in Fig.~\ref{fig:gauss}.

\textbf{Weather Robustness.} We evaluate the framework under adverse conditions in Table~\ref{tab:weather}. In Rainy and Night scenarios, where camera data is often degraded, our model achieves 27.1\% and 15.9\% mIoU, respectively. This performance significantly exceeds OccFusion \cite{occfusionming2024} and GaussianFormer3D \cite{gaussianformer3d2025}, proving that our  fusion effectively leverages LiDAR geometry to compensate for visual signal loss.

\vspace{-0.2cm}
\begin{table}[h]
\centering
\scalebox{0.85}{
\begin{tabular}{c|c|c|c}
\toprule
Method  &  Modality & Number of Params. (M) & Latency (ms)         \\ 
\midrule
       GaussianFormer~\cite{huang2024gaussianformer} & C & \textbf{52.4}  & \textbf{342}  \\ 
       \midrule
      OccMamba~\cite{occmamba2024}& C\&L &92.3 & 531  \\
      \rowcolor{black!10}
      \textbf{GaussianOcc3D (ours)} & C\&L & \underline{68.1} & \underline{427} \\ 
\bottomrule
\end{tabular}
}
\vspace{-5pt}
\caption{Analysis of different model complexity. The best and second-best are in bold and underlined, respectively.}
\label{table:complexity}
\end{table}

\vspace{0.15cm}
\textbf{Component Analysis.} As shown in Table~\ref{table:abb}, we investigate the impact of our proposed modules. Baseline performance starts at 20.2\% mIoU. The inclusion of LiDAR data provides a jump to 24.1\%, while the addition of our adaptive fusion and \textbf{LDFA} strategy further elevates the performance to 28.4\%. Finally, incorporating \textbf{EBFS} and the \textbf{Gauss-Mamba Head} allows the model to reach its peak performance of 28.9\% mIoU, confirming that each module contributes uniquely to spatial awareness and feature refinement.

\begin{table}[htbp]
\centering
\resizebox{\linewidth}{!}{ 
\begin{tabular}{ccccc|c}
\toprule
LiDAR & ACLF & LDFA & EBFS & Gauss-Mamba head & mIoU \\ 
\midrule
      &            &      &           &            & 20.2 \\ 
\checkmark &       &      &           &            & 24.1 \\ 
\checkmark & \checkmark &      &           &            & 26.1 \\ 
\checkmark & \checkmark & \checkmark &      &           & 28.4 \\
\checkmark & \checkmark & \checkmark & \checkmark &      & \underline{28.6} \\
\rowcolor{black!10}
\checkmark & \checkmark & \checkmark & \checkmark & \checkmark & \textbf{28.9} \\
\bottomrule
\end{tabular}
}
\vspace{-5pt}
\caption{Ablation study of different components on SurroundOcc \cite{surroundocc} validation set. The best and second-best are in bold and underlined, respectively.}
\label{table:abb}
\end{table}

\vspace{-0.2cm}
\section{Conclusion}
\label{sec:conclusion}

In this paper, we proposed a multi-modal 3D semantic occupancy prediction framework that leverages 3D Gaussian primitives to bridge the gap between sparse geometric signals and dense volumetric representations. Our approach improves the interaction between LiDAR and camera data through the adaptive camera-LiDAR fusion module and a novel LiDAR depth feature aggregation strategy. We introduced entropy-based feature smoothing to refine the semantic distribution of the primitives, ensuring clearer boundaries and reduced noise in the occupancy grid. Furthermore, we incorporated the Gauss-Mamba Head to capture global spatial dependencies with linear complexity, facilitating a more comprehensive understanding of complex driving scenes. Extensive experiments on the NuScenes and SemanticKITTI benchmarks confirm the effectiveness of our proposed framework in generating high-fidelity 3D semantic occupancy predictions.

\IEEEpeerreviewmaketitle

{\small
\bibliographystyle{ieeetr}
\bibliography{bibtex/ref}}

\end{document}